\documentclass[letterpaper, 10 pt, conference]{ieeeconf}  

\IEEEoverridecommandlockouts                              
\usepackage{cite}
\usepackage{amsmath} 
\usepackage{amssymb}  
\usepackage{stackengine}
\stackMath
\usepackage{cite}
\usepackage[ruled,noend]{algorithm2e}
\usepackage{graphicx}
\usepackage{caption}
\usepackage{subcaption}
\usepackage{hyperref}
\usepackage[switch]{lineno}
\usepackage{mathtools}

\makeatletter
\renewenvironment{subarray}[2][c]{%
  \if#1c\vcenter\else\vbox\fi\bgroup
  \Let@ \restore@math@cr \default@tag
  \baselineskip\fontdimen10 \scriptfont\tw@
  \advance\baselineskip\fontdimen12 \scriptfont\tw@
  \lineskip\thr@@\fontdimen8 \scriptfont\thr@@
  \lineskiplimit\lineskip
  \ialign\bgroup\ifx c#2\hfil\fi
    $\m@th\scriptstyle##$\hfil\crcr
}{%
  \crcr\egroup\egroup
}
\makeatother

\usepackage{xcolor}
\usepackage{graphicx}
\usepackage{soul}

\newcommand{\Tau}{\mathcal{T}}

\DeclareMathOperator*{\argmin}{arg\,min}

\title{\LARGE \bf
Teaching Robots to Span the Space of Functional Expressive Motion
}

\author{Arjun Sripathy$^{1}$, Andreea Bobu$^{1}$, Zhongyu Li$^{1}$, Koushil Sreenath$^{1}$, Daniel S. Brown$^{2}$, and Anca D. Dragan$^{1}$
\thanks{*This research is supported by the Office of Naval Research (ONR-YIP), the Air Force Office of Scientific Research (AFOSR), the UCSF Weill Institute for Neuroscience, and the Apple AI/ML PhD Fellowship. $^{1}$UC Berkeley {\tt\small \{arjunsripathy, abobu, zhongyu\_li, koushils, anca\}@berkeley.edu}. 
$^{2}$ University of Utah {\tt\small dsbrown@cs.utah.edu}
}
}

\begin{document}

\newcommand{\traj}{\ensuremath{\xi}}
\newcommand{\task}{\ensuremath{\tau}}
\newcommand{\costb}{\ensuremath{C_{base}}}
\newcommand{\costs}{\ensuremath{C_{style}}}
\newcommand{\style}{\ensuremath{e}}
\newcommand{\weight}{\ensuremath{\theta}}
\newcommand{\embedding}{\ensuremath{f_{\weight}}}
\newcommand{\hlabel}{\ensuremath{l}}
\newcommand{\equery}{\ensuremath{s}}
\newcommand{\tquery}{\ensuremath{q}}
\newcommand{\rounds}{\ensuremath{K}}
\newcommand{\batches}{\ensuremath{B}}
\newcommand{\round}{\ensuremath{k}}

\newcommand{\trajset}{\ensuremath{\Xi}}
\newcommand{\taskset}{\ensuremath{\Tau}}
\newcommand{\styleset}{\ensuremath{\mathcal{E}}}
\newcommand{\tqueryset}{\ensuremath{\mathcal{Q}}}
\newcommand{\equeryset}{\ensuremath{\mathcal{S}}}
\newcommand{\hlabelset}{\ensuremath{\mathcal{L}}}

\maketitle
\thispagestyle{empty}
\pagestyle{empty}

\begin{abstract}
Our goal is to enable robots to perform functional tasks in emotive ways, be it in response to their users' emotional states, or expressive of their confidence levels. Prior work has proposed learning independent cost functions from user feedback for each target emotion, so that the robot may optimize it alongside task and environment specific objectives for any situation it encounters. However, this approach is inefficient when modeling multiple emotions and unable to generalize to new ones. In this work, we leverage the fact that emotions are not independent of each other: they are related through a latent space of Valence-Arousal-Dominance (VAD). Our key idea is to learn a model for how trajectories map onto VAD with user labels. Considering the distance between a trajectory's mapping and a target VAD allows this single model to represent cost functions for all emotions. As a result 1) all user feedback can contribute to learning about every emotion; 2) the robot can generate trajectories for any emotion in the space instead of only a few predefined ones; and 3) the robot can respond emotively to user-generated natural language by mapping it to a target VAD. We introduce a method that interactively learns to map trajectories to this latent space and test it in simulation and in a user study. In experiments, we use a simple vacuum robot as well as the Cassie biped.

\end{abstract}


\section{Introduction}
\label{sec:intro}



\begin{figure}
    \includegraphics[width=0.49\textwidth]{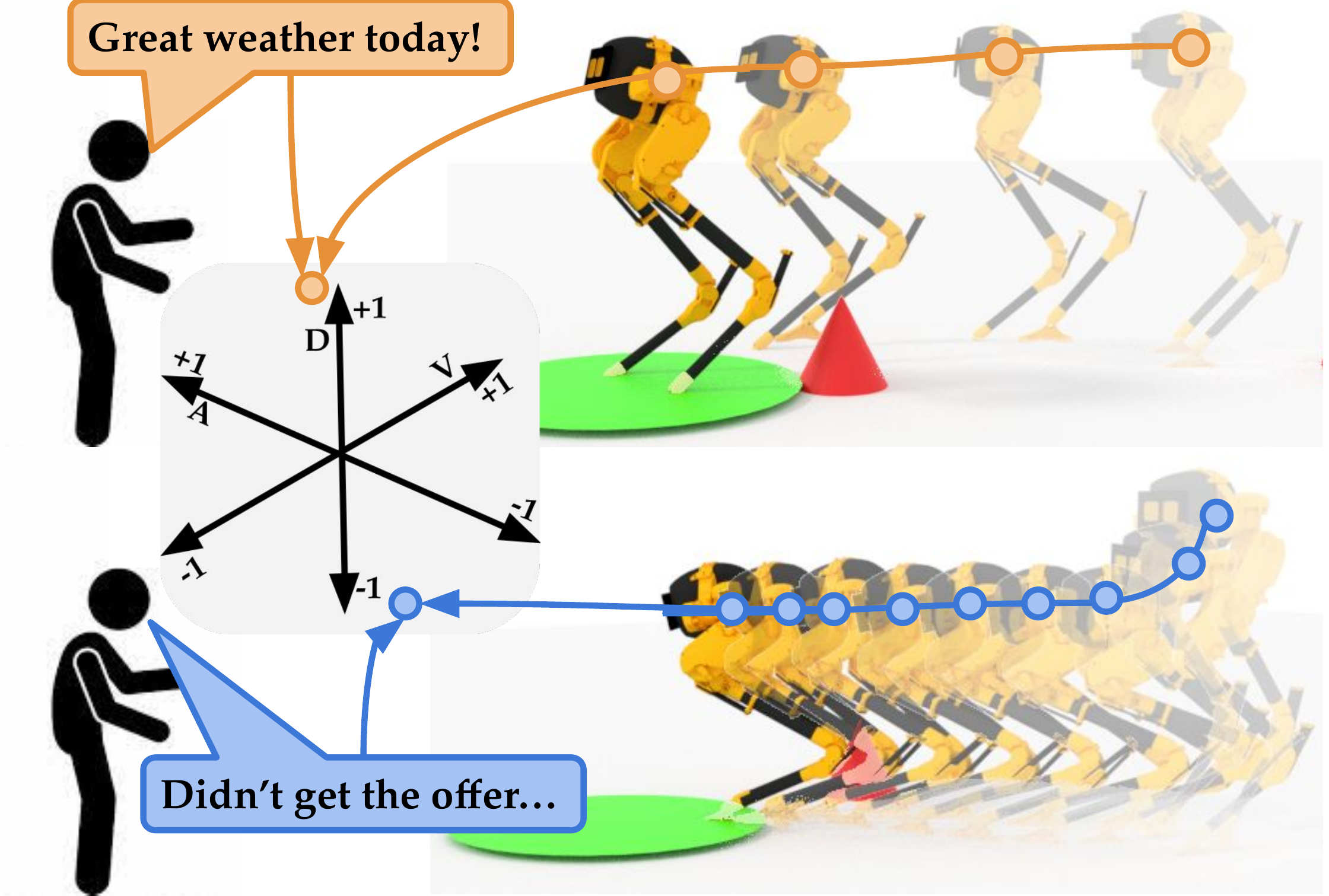}
    \centering
    \caption{Cassie robot performing a task with trajectories it believes exhibit similar emotion VAD as the human speech. (Top) The person's happier sentiment (orange) results in the robot's fast upright motion. (Bottom) The person's sad sentiment (blue) results in slow and slouched motion.}
    \vskip -12pt
    \label{fig:front_fig}
\end{figure}

Robotics research tends to focus on generating functional motion, in service of the robot's task. But imagine coming home from work exhausted and disappointed in being rejected from a job application, and the robot continuing to carry on its tasks as if nothing changed. Or, coming back with high energy after taking a walk in the sun, and the robot still putting the dishes away in the same exact way it always does. Ideally, our robots should adapt their behavior like in Fig.~\ref{fig:front_fig}, including the way they move, in response to us, as well as in response to successes and failures they encounter, their confidence levels when performing a task, etc.

While much work has focused on expressive or emotive robot \emph{gestures} \cite{Suguitan2020DemonstratingMM, Desai2019GeppettoES, Saerbeck2010PerceptionOA}, the ability to generate emotive \emph{functional} motion that still achieves the robot's task remains an open problem. How can a robot walk to its goal and avoid obstacles \emph{while seeming happy or confident} in response to its user having had a great day? How could a manipulator place a dish in the sink while empathizing with its user's disappointment at work? Taking an existing motion and adjusting its affective features, as researchers do with gestures~\cite{gielniak2011anticipation,Suguitan2020DemonstratingMM}, would no longer meet the functional task specification. Instead, prior work \cite{Zhou2018CostFF} has proposed to learn a \emph{cost function} from user feedback for each desired emotion or style, that can be then optimized along with the task specification. Although this addresses the problem of generating motion in an emotive style even as the specifics of the task change, it has the challenge that we have to think of every desired emotion in advance, and collect data specific to it. Further, we still need a way to decide which style or emotion to use.

In our work, we focus on the fact that emotions are not independent---they are latently related through the Valence-Arousal-Dominance (VAD) spectrum. Motivated by foundational studies in social psychology, VAD identifies three continuous, interpretable directions capturing much of emotional variance \cite{osgood1957, russell1980}.
Rather than learning independent cost functions for each emotion, our key idea is to learn to map robot trajectories to an emotive VAD latent space---this way, all user feedback contributes to learning about all emotions, and the robot can model new emotions that interpolate those seen during training. This enables robots to perform tasks in ways expressive of any specific emotion, by optimizing for a trajectory with a projection onto the latent space that is as close as possible to the desired emotion's VAD. They also may use natural language to infer target emotion VAD: enabling stylistic response to emotive words, like ``anger'', or even sentences, like ``Great weather today!'' as in Fig.~\ref{fig:front_fig}.

Our approach interactively collects data from a user to learn this emotive latent space: it starts with an initial space, uses it to optimize emotive trajectories for a variety of task specifications and target emotions, asks the user to label these trajectories, and retrains the latent space to agree with the user labels. Users may choose to label directly with VAD, or use language, which we can map to VAD by using pre-trained language models~\cite{Devlin2019BERTPO} finetuned to predict VAD scores.

In experiments with simulated human feedback for a Vacuum robot and the Cassie biped, we demonstrate the efficiency of our method in learning emotive costs when compared to approaches which model each emotion independently. 
We then show in a user study with the Vacuum robot that real humans can teach personalized emotive style in only 30 minutes of labeling. 
We find that users are able to recognize target emotions in robot motions generated with the model trained on their labels, even though those target emotions were not explicitly queried for during training.
In summary, we propose a method for generating functional, stylistic robot motion by efficiently teaching how trajectories map to VAD defining a cost function encouraging target emotive style specified by natural language. Code and videos are made available at \href{https://arjunsripathy.github.io/robot_emotive_space}{arjunsripathy.github.io/robot\_emotive\_space}.

Despite showing promise in enabling simulated robots to perform functional tasks while expressing a wide range of emotions, much work remains ahead. Demonstrating generalization to a broader range of tasks than locomotion, thinking critically about how target VAD should be determined based on the emotion the user expresses, and moving as much of the process as possible to the physical domain all pose interesting challenges which we discuss further in Sec.~\ref{sec:limits}

\section{Related Work}
\label{sec:related_work}

Getting robots and virtual avatars to exhibit realistic looking and human-recognizable motions is a well-studied problem, from conveying intent in a task~\cite{dragan2013legibility, gielniak2011anticipation,szafir2014flyers}, to communicating incapability~\cite{kwon2018incapability, takayama2011expressingthought}, to expressing emotions~\cite{knight2014expressive,sharma2013locomotion,Zhou2018CostFF}.
In this section, we focus our attention on literature from the latter category, as our goal is enabling robots to learn emotive styles for performing functional tasks.

Motion style research has its roots in the graphics community. 
Some work looks at transferring motion capture style from one clip to another~\cite{torresani2006learning,xia2015styletransfer,holden2016charactermotion}, but such unconstrained transfer is not appropriate for robots that need to satisfy rigid physical dynamics,
 or, even more challenging, to still be performing the desired underlying task.
Alternative approaches use human demonstrations to learn locomotion styles as cost functions that the robot optimizes to respect task constraints~\cite{liu2005style,lee2010style}.
Unfortunately, due to the correspondence problem in robotics~\cite{ARGALL2009lfdsurvey}, these methods cannot be applied outside of locomotion robots, and acquiring demonstrations of stylized non-anthropomorphic robots is extremely challenging, especially when moving beyond gestures to functional motion.

In typical robotics motion style work, researchers design libraries of emotive motions that the robot can use during task execution~\cite{Lim1999EmBiped,knight2014expressive,sharma2013locomotion,Li2020AnimatedCA}. 
To produce trajectories feasible for complex physical systems, Li et. al.~\cite{Li2020AnimatedCA} employ a dynamically constrained optimization that encourages the resulting motion to match stylized trajectories while abiding by the robot's dynamics. 
The motions in these methods are hand-crafted and, therefore, specific to the system and task they are being designed for.
To generalize to a more diverse set of tasks, recent methods~\cite{Zhou2018CostFF,liu2005style,lee2010style} try to learn a cost function that when optimized produces the desired emotive motion. However, these methods requires collecting labels for each emotion one at a time, resulting in inefficient and costly learning that fails to generalize to new emotions.

Instead of representing each emotive motion with an individual cost function, we can learn a more generalizable representation. Suguitan et. al.~\cite{Suguitan2020DemonstratingMM} learned a latent emotive embedding along the VA spectrum \cite{osgood1957, russell1980} capturing a whole space of emotions. While their approach enabled the robot to exhibit simple emotive gestures, like a slow lowering of the head for \textit{sad}, we are interested in integrating emotive motion during the robot's task execution.
In this work, we take inspiration to similarly learn an embedding that maps emotive trajectories like the ones in Fig. \ref{fig:front_fig} to a latent VAD space but extend their approach to functional task behaviors. 

With this embedding, we have a representation that is both generalizable to new emotions and amenable to alternative forms of human feedback, such as natural language. Recently large, pre-trained language models such as BERT~\cite{Devlin2019BERTPO} have made transfer learning for downstream natural language tasks more accessible and efficient. Further, due to the breadth of research around VAD there exist datasets containing language and corresponding manual VAD annotations \cite{vad-acl2018, Buechel2017EmoBankST}. Putting these together we may train a model for mapping natural language into our learned VAD space which will allow us to make the interaction between user and our system even more seamless.
\section{Method}

\begin{figure}
    \includegraphics[width=0.49\textwidth]{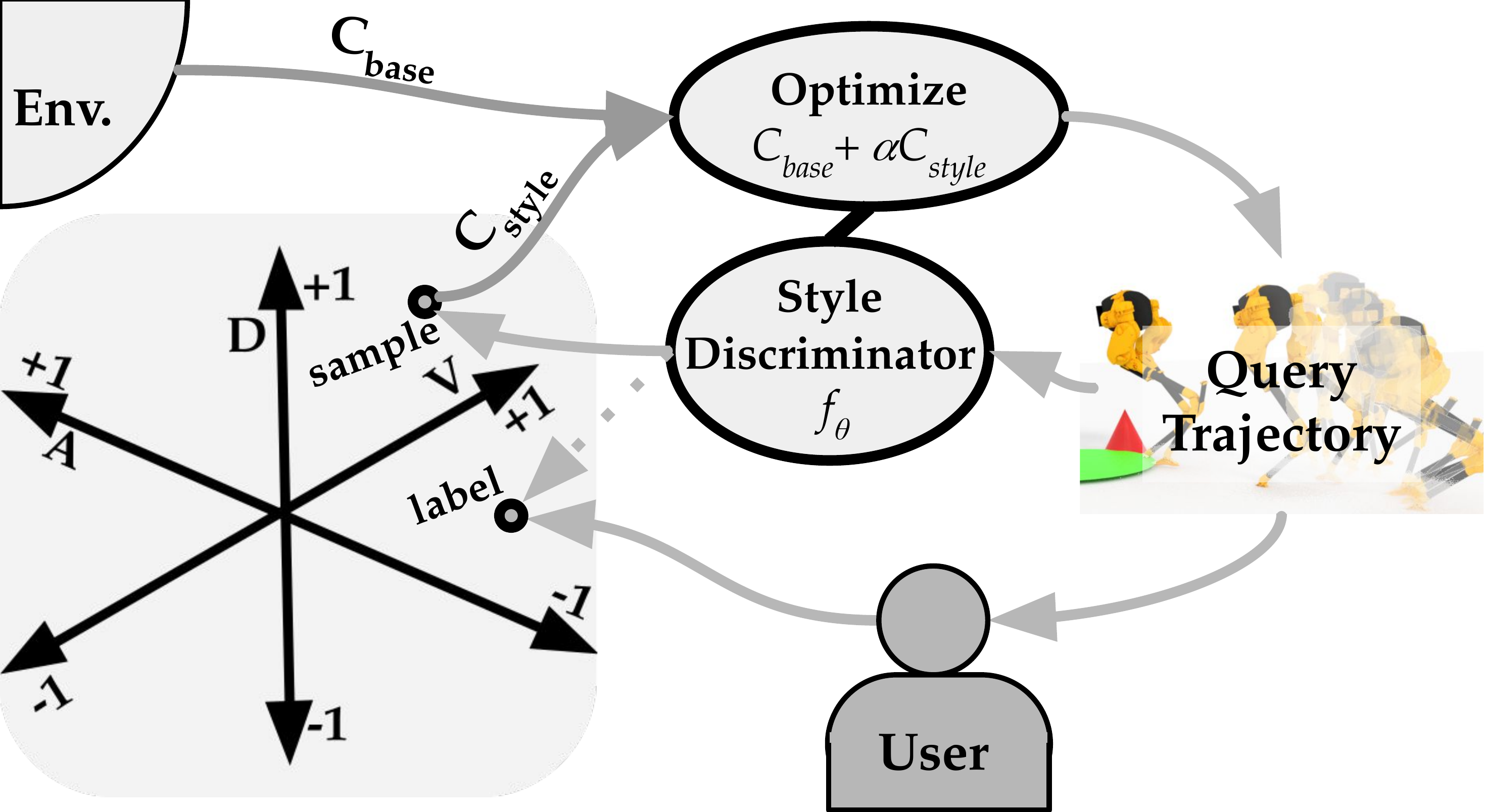}
    \centering
    \caption{Visualization of our method. The optimizer combines the task's base cost $\costb$ with the style cost $\costs$ of a sampled emotion to produce a query trajectory the style discriminator believes aligns with the target VAD. The user labels the trajectory with a VAD, and the style discriminator $\embedding$ is trained to bring its predicted VAD closer to the human label.}
    \vskip -12pt
    \label{fig:method_fig}
\end{figure}

We now present our method for enabling a human to teach robots how to exhibit a broad range of emotions while performing various tasks. The core of our method is training a style discriminator, which predicts what emotion the human would perceive given a trajectory, using VAD labels collected from humans in response to query motions. For any target emotion, we'll define a motion style cost based convincing the discriminator that the trajectory being optimized exhibits the target VAD. We leverage the interpretable structure of VAD as a representation for emotion to improve learning efficiency, interpolate or extrapolate to new emotions, and integrate natural language seamlessly.


\subsection{Preliminaries}
\label{sec:prelims}
We represent a trajectory $\traj \in \trajset$, where $\trajset$ denotes the set of all trajectories in an environment, as a variable length sequence of waypoints along with the variable time duration between each pair of consecutive points. We denote environment tasks, such as moving from a specific start location to a specific goal, as $\task \in \taskset$, where $\taskset$ denotes the set of all tasks in the environment.
The robot produces a trajectory that solves a particular task in the environment by optimizing a \textit{base cost} $\costb: \trajset \times \taskset \rightarrow \mathbb{R}$.
Optimizing a trajectory using $\costb$ yields an efficient trajectory but offers no control over the emotion and typically produces a neutral style.

We describe the style of a trajectory based on an emotion VAD latent $\style \in \styleset \coloneqq [-1, 1]^3$, where the three values continously represent Valence, Arousal, and Dominance in that order. Our goal is to learn a trajectory \textit{style cost}, $\costs: \trajset \times \styleset \rightarrow \mathbb{R}$, capable of encouraging stylistic alignment with any target emotion $\style$.  Ultimately, to produce trajectories that achieve the task with the target style the robot will trade off between the base cost and the style cost:
\begin{equation}
    C(\traj, \task, \style) = \costb(\traj, \task) + \alpha \cdot \costs(\traj, \style) \enspace ,
    \label{eq:ovr_cost}
\end{equation}
where $\alpha$ is a user specified hyperparameter that prioritizes between style and efficiency.

\subsection{Cost Function Formulation}\label{subsec:cost_fn_form}
To learn the style cost $\costs$, we propose training a neural network style discriminator $\embedding: \trajset \rightarrow \styleset$ parameterized by $\theta$ to map a trajectory $\traj$ to the emotion $\style$ the robot exhibits while following it. Our motivation for this design is that every trajectory exhibits some emotion. The style discriminator $\embedding$ naturally motivates a style cost function $\costs$ which penalizes a trajectory $\traj$ based on how much its exhibited emotion, $\embedding(\traj)$,  differs from the target $\style$. We formalize this intuition using Euclidean distance in $\styleset$:
\begin{equation}
    \costs(\traj, \style) = ||\embedding(\traj) - \style||_2^2\enspace.
    \label{eq:style_cost}
\end{equation}
By optimizing the combined cost function in Eq.~\eqref{eq:ovr_cost} along with any task constraints, the resulting trajectory $\traj$ completes the task while making its best effort to exhibit the target $\style$. 

\subsection{Generating Human Queries}
\label{sec:query}

To train a robust discriminator $\embedding$, we generate batches of trajectories and query the user for emotive labels as shown in Fig.~\ref{fig:method_fig}. The user provides either direct VAD labels or language which we map to VAD as discussed in Sec.~\ref{sec:lang_vad}. Our goal is to learn $\theta$, which is randomly initialized and updated after each labeling round as discussed in Sec.~\ref{section:training_proc}.

To generate a round of query trajectories we optimize Eq.~\eqref{eq:ovr_cost} for a batch of sample emotions and tasks, using the current estimate of $\theta$ for $\costs$. Given $\costs$ does not explicitly model the task, we focus on how to sample emotions in a way that is most informative for $\theta$. Motivated by active learning literature we seek a diverse batch that biases towards \textit{important}, unexplored areas of $\styleset$ \cite{Zhdanov2019DiverseMA}.


\begin{figure}
    \includegraphics[width=0.49\textwidth]{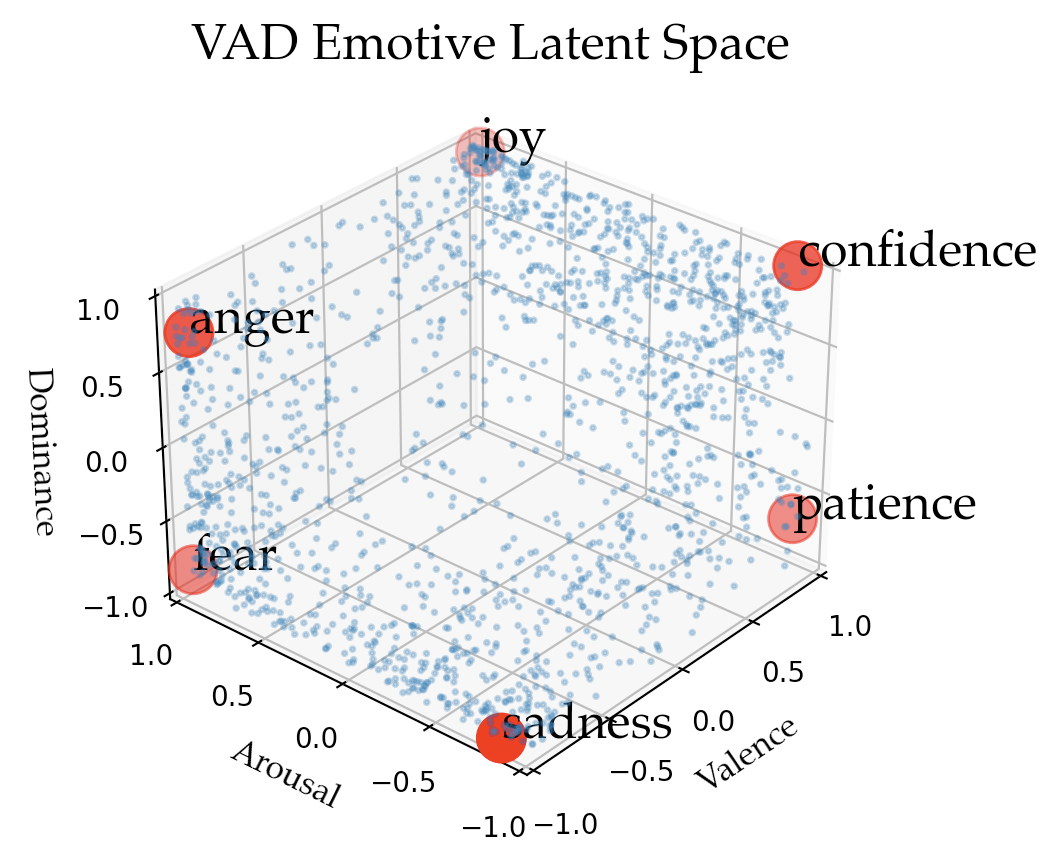}
    \centering
    \caption{The VAD latent space with each of Valence, Arousal, and Dominance being a real valued axis ranging from -1 to 1.  The scatter plot depicts the projections of 1,672 emotive words projections into this space with the red dots highlighting the 6 basic evaluation emotions we used in our experiments.}
    \vskip -12pt
    \label{fig:vad_visual}
\end{figure}

Not all areas of $\styleset$ are equally important. Emotions are not uniformly spread across $\styleset$, and we would like to focus our queries on more populated areas of the space. 
We leverage the empirical emotion distribution from the NRC VAD lexicon~\cite{vad-acl2018}, which contains annotated VAD values for 20k words.
We filtered them down to 1,672 common emotive words, resulting in the VAD distribution in Fig.~\ref{fig:vad_visual}.

We now propose an active learning method for improving query coverage of this distribution to make the discriminator, $\embedding$, more robust. For the first round of queries, since the network $\embedding$ is randomly initialized and has no semantic meaning yet, we uniformly sample $\batches$ emotions from $\styleset$.
To explain the process for successive rounds we must establish some notation. We conduct $K$ query rounds with $k \in [1, K]$ referencing the round index. Let $\equeryset_k \coloneqq \{\equery_k\}^{1:\batches} \in \styleset^\batches$ reference the batch of sample emotions to be chosen by active learning to cover the empirical distribution of $D=~1672$ lexicon VAD values $\style_1, ..., \style_D$. By optimizing Eq.~\ref{eq:ovr_cost} for $\equeryset_k$ alongside tasks randomly presented by environment, we generate query trajectories $\tqueryset_k \coloneqq \{\tquery_k\}^{1:\batches} \in \trajset^\batches$. For these queries we will collect human labels referenced $\hlabelset_k~\coloneqq~\{\hlabel_k\}^{1:\batches} \in \styleset^\batches$.

Our active learning method seeks to minimize the average distance between lexicon emotions and the closest acquired label from any round. This relies on estimating the $\hlabelset_k$ based on our selection of $\equeryset_k$. Our approximation here is $\hlabelset_k \approx \equeryset_k$ which becomes more accurate over the course of training. With this assumption, for $k>1$, we may select  $\equeryset_k$ as: 
\begin{equation}
    \equeryset_k = \argmin_{\equeryset_k}{\sum_{i=1}^{D} \min_{l \in \equeryset_k \cup \bigcup\limits_{j=1}^{\round-1} \hlabelset_j}  ||\style_i - l||^2}\enspace.
    \label{eq:query}
\end{equation}

$\equeryset_k$ will bias towards densely populated areas of the VAD space where we do not yet have a nearby label. We approximate the optimal solution using expectation maximization\cite{Dellaert2010ExpectationMaximizationA}. We now turn our attention to how we may update $\theta$ after each round based on the collected human feedback.

\subsection{Trajectory Network Training and Architecture}
\label{section:training_proc}
After each round $\round$ of querying, we update $\theta$ given our trajectory queries and VAD label responses collected so far ($\tqueryset_{1:\round}$, $\hlabelset_{1:\round}$). We optimize the following MSE training loss:
\begin{equation}
    L_\round(\theta, \tqueryset_{1:\round}, \hlabelset_{1:\round}) = \sum_{i=1}^{\round} \sum_{j=1}^{\batches} ||\embedding(\tquery_i^j) - \hlabel_i^j||^2_2\enspace.
    \label{eq:loss}
\end{equation}
Note the summand is exactly equivalent to $\costs(\tquery_i^j, \hlabel_i^j)$ allowing for an alternative interpretation: we treat the queries as demonstrations for the emotion labels and would like to assign them minimal style cost.

In implementing this method, we have to choose a specific architecture for $\embedding$. Recall from Sec.~\ref{sec:prelims} that we represent a trajectory as a variable-length sequence of waypoints and time deltas. We utilize an architecture similar to PointNet \cite{Qi2017PointNetDL} for its simplicity and ability to gracefully handle varying length trajectories. First, a fully connected network processes each waypoint independently. Then we apply average and softmax pooling over waypoints to produce a single trajectory embedding. From there another fully connected network predicts the overall trajectory VAD value. Both networks use ELU activation \cite{Clevert2016FastAA}.

It is important that the network predictions are smooth so they guide trajectory optimization well when used within the cost function. In other words, not only must predictions be accurate, but their gradient signal must also be informative. These factors motivated us to use smoother pooling (average \& softmax) and activation functions (ELU), and to limit network capacity. A single hidden layer in each network, of dimensions chosen to match the complexity of the robot, along with L1 regularization worked well in our experiments.

\subsection{Natural Language to VAD}
\label{sec:lang_vad}
We now describe how VAD may be inferred from natural language and where this may be used. Single words we look up directly in the NRC VAD lexicon \cite{vad-acl2018}. For sentences, or words not present in the lexicon, we apply a BERT model finetuned to predict VAD using EmoBank: a dataset with 10k VAD labeled sentences \cite{Devlin2019BERTPO, Buechel2017EmoBankST}. The wealth of resources and data around VAD is another benefit of using the spectrum as our latent space.

In many scenarios language provides a more natural means of communicating emotion than VAD. During training, language labels could be easier to provide compared to VAD directly. To be practical after training, the robot likely must determine target emotion in a less burdensome way then explicitly requesting VAD. Interpreting VAD from language allows the robot to seamlessly identify target emotion and modulate its behavior around humans.
\begin{figure*}
    \includegraphics[width=0.99\textwidth]{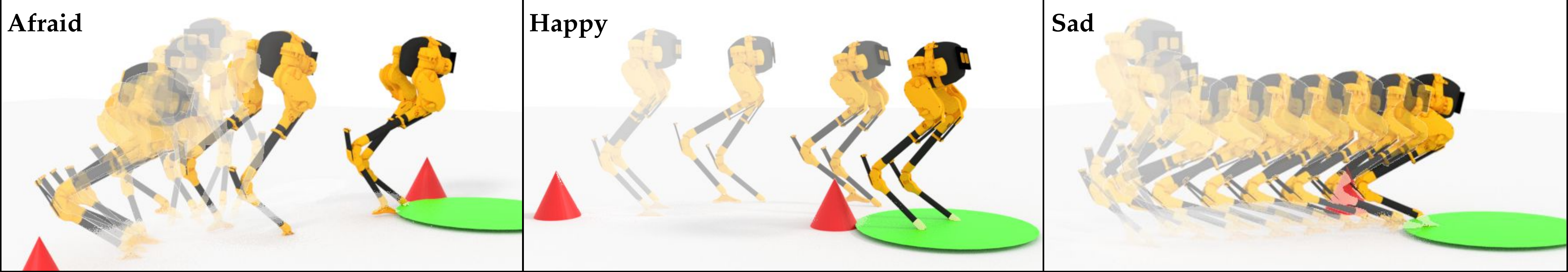}
    \centering
    \caption{Cassie must reach a target location avoiding obstacles represented by cones.  Here it optimizes for various target emotions on various tasks using a model learned with our method.  Visuals overlay one snapshot per second with earlier frames made more transparent. \emph{Afraid} first takes a few cautious steps back before proceeding. \emph{Happy} keeps head high and moves fast. \emph{Sad} slouches and proceeds slowly.}
    \vskip -12pt
    \label{fig:cassie_visual}
\end{figure*}

\section{Experiments}
\label{sec:experiments}

There are three primary hypotheses we seek to test with our experiments. (1) A real human is capable of using our method to teach their perception of a robot's emotive style, and after training they perceive the robot's intended target emotion in generated stylistic trajectories; (2) Our method is more efficient at learning a set of emotive styles than an approach that models each emotion independently; (3) Despite using general emotive labels, our method is equally efficient at learning any single emotive style as alternatives which leverage feedback specific to that emotion.

To evaluate hypothesis (2) \& (3) we run a set of simulated human trials comparing the query efficiency our method to alternatives which model emotions independently. To test (1) we conduct a user study where real humans' perception of emotion takes the place of simulated human heuristics. We evaluate the effectiveness of teaching by the extent to which the human (simulated or real) perceives emotion similar to what the robot intended to exhibit while completing various tasks. We will discuss the results for each case and ultimately find our hypotheses supported by the data collected.

\subsection{Robots}
We used two simulated robots to experiment with our method: a simpler Vacuum robot (VacuumBot) and the more complex and realistic Cassie bipedal robot \cite{Li2020AnimatedCA}. We now describe the robot specifications, the environments they operated in, and the tasks they must complete.

VacuumBot is tasked with moving to collect dust that appears in a 2D world. It has 3 DOF controlling horizontal, vertical and angular acceleration and is subject to various physical constraints including gravity and friction. The current state of the robot and environment is summarized by the position and velocity corresponding to the DOF as well as the location of the dust. Trajectories are optimized for VacuumBot entirely using PyTorch \cite{Paszke2019PyTorchAI}.

Cassie, shown in Fig.~\ref{fig:cassie_visual}, is a person-sized bipedal robot which has 20 DoF including 6 DoF of the base (its pelvis), 5 DoF and 2 passive joints of each leg. More details about Cassie can be found in~\cite{Li2020AnimatedCA}.
In our paper, Cassie is tasked with navigating to a randomly generated target location while avoiding random obstacles represented by the red cones in Fig.~\ref{fig:cassie_visual}. 
We leverage collocation to obtain a trajectory minimizing the proposed style cost while imposing constraints including collision-avoidance and reduced-order nonlinear dynamics as described in~\cite{li2021vision}.
Based on the optimized trajectory of robot base velocity and height, the robot's whole body motion is obtained from a gait library optimized by its full-order dynamics~\cite{Li2020AnimatedCA,hereid2019rapid}.
This nonlinear optimization is formulated in CasADI~\cite{andersson2019casadi}, solved via IPOPT~\cite{biegler2009large}, and the resulting trajectory is visualized through animation in Blender~\cite{Li2020AnimatedCA}. To integrate our PyTorch implementation of $\embedding$ we export our learned $\theta$ after every training round and replicate the neural network as a fixed numerical function in CasADI~\cite{andersson2019casadi}.

\subsection{Experimental Design}
The simulated human trials and real human study used a very similar experimental design. In this section, we describe the process for evaluating a single method in general terms, and will discuss how this was adapted for the two contexts in their corresponding sections. We implement alternative models in a way that allows them to conform to the same evaluation procedure as our method. 

Using notation from Sec. \ref{section:training_proc}, we train by conducting $K$ rounds of $\batches$ trajectory query batches updating the model after each batch of labels. Then we evaluate stylistic trajectories, produced for a representative set of evaluation emotions, based on the extent to which the human perceives the intended emotion in each.

To identify our evaluation emotions, we again leveraged VAD values from the NRC VAD Lexicon \cite{vad-acl2018}. Based on the empirical distribution shown in Fig. \ref{fig:vad_visual}, we identified the corners [-1, -1, -1], [1, 1, 1], [-1, 1, -1], [1, -1, 1], and [-1, 1, 1] as the best regions to evaluate our model due to their population density and general coverage of the space. We selected representative emotions near each of these corners: \emph{sadness}, \emph{joy}, \emph{fear}, \emph{confidence}, \emph{anger}, and \emph{patience} respectively. Note that consecutive pairs on this list are diametric opposites in VAD space. We will not always use all 6 evaluation emotions and use $N$ to reference the specific number we are working with. For consistency, when $N < 6$ we will always use the \textit{first} $N$ emotions based on the order we presented these above. We restrict ourselves to $N \in \{2, 4, 6\}$ to keep the emotions in opposing pairs.

Our three evaluation metrics are \textit{Quality} score, \textit{Top-1} accuracy, and \textit{Top-2} accuracy. Quality score measures binary alignment: how well trajectories express the intended emotion compared to its diametric opposite. Top-X accuracies measure precise alignment: how well trajectories express the intended emotion compared to all $N - 1$ alternatives. Ideally we'd evaluate across all tasks, but given the task space is continuous we randomly sample $M$ tasks for each of the $N$ evaluation emotions averaging the metric values we get.

To compute the quality score, the robot presents the user with $N/2$ sets of $2\cdot M$ trajectories for evaluation. Each set is associated with one diametric pair of evaluation emotions, say A and B. The $2\cdot M$ trajectories includes $M$ trajectories optimized for A and $M$ trajectories optimized for B. The user is asked to assign each trajectory a score, $s$ from 1 to 7 answering the Likert question: \emph{Is the trajectory more expressive of Emotion B than A?} with a response of 1 indicating the trajectory is very expressive of Emotion A and a 7 indicating that the trajectory is very expressive of B. Let $q$ be the Quality score metric. For trajectories optimized for B we define $q \coloneqq s$, and for A $q \coloneqq 8 - s$. As a result, $q$ ranges from 1 to 7 with 7 indicating perfect alignment with the intended emotion compared to its opposite.

To compute Top-X accuracies, we present the user with another $N\cdot M$ trajectories including $M$ trajectories optimized for each of the evaluation emotions. Now the user is asked to select which of the $N$ emotions is most expressed by each trajectory as well as their second choice. We define the Top-X accuracy metric, for $X \in \{1, 2\}$, as the proportion of the time the user's top $X$ choices include the intended emotion.

\subsection{Simulated Human Trials}
\label{sec:sims}
We first conducted a set of experiments with simulated human feedback, since it would have been impractical to reliably test all our configurations with real humans.

\subsubsection{Simulating Human Feedback}
The simulated ``human'' (SH) uses heuristics to determine VAD for trajectories. For example, SH quantifies dominance for Cassie based on the average head height. SH may not accurately represent human emotive perception, but its consistent feedback allows us to compare learning efficiency between various methods.

During training SH directly provides its determined VAD for a trajectory as feedback. During evaluation, it must further transform this VAD value to mimic appropriate human responses. For the Likert question juxtaposing opposite emotions, SH projects the VAD value on to the diametric axis between the pair of emotions; then SH linearly scales the result so exactly Emotion A is 1 and exactly B is 7 clipping outside that range. For the choice based component, SH picks the closest and second closest evaluation emotion based on Euclidean distance in VAD space.

\begin{figure*}
    \includegraphics[width=0.99\textwidth]{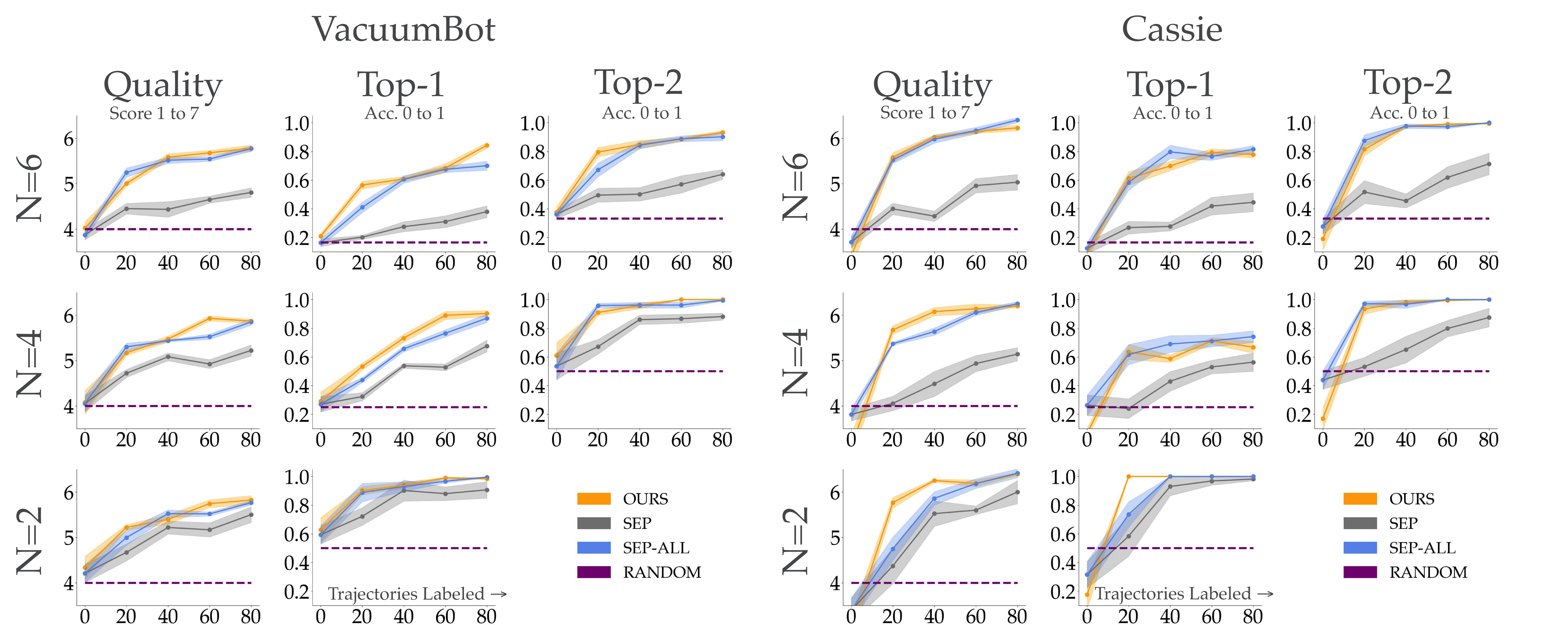}
    \centering
    \caption{Quality score, Top-1, Top-2 accuracy and standard errors over the course of training averaged across six seeds for each (environment, method, $N$) configuration. Metrics were computed before training and after each of 4 batches of 20 trajectory queries, with the query number indicated in the horizontal axis. Shown in dotted lines are the expected values if users chose randomly during evaluation. Our method is able to match performance of SEP-ALL whereas SEP struggles to keep pace as $N$, the number of evaluation emotions, increases.}
    \vskip -12pt
    \label{fig:sim_evals}
\end{figure*}

\subsubsection{Alternative Methods}
We compare our method to two alternative approaches which model emotions independently to test hypotheses (2) \& (3). First is an approach inspired by Zhou et al. \cite{Zhou2018CostFF} which we'll reference as SEP, because it trains \textit{separate} models directly predicting cost for each evaluation emotion. To do so, SEP directly asks the user to label trajectories with how expressive they are of \emph{one} of the evaluation emotions. The second approach is SEP-ALL which we allow access to real valued cost labels for \emph{all} evaluation emotions for each trajectory; SEP-ALL does not have to split its labeling budget between emotions as SEP does. Recall our approach requests VAD labels from the user irrespective of the evaluation set enabling generalization beyond predefined emotions. In contrast, SEP and SEP-ALL both require knowledge of the evaluation set of emotions prior to training and get feedback specific to them.

To select emotions to generate trajectory queries for, SEP and SEP-ALL simply sample with replacement from the evaluation set. To remove a potential confound and isolate learning efficiency, we use the same selection process for our method in the simulated experiments. In the Sec.~\ref{sec:study} study we revert to the active learning described in Sec.~\ref{sec:query}.

\subsubsection{Simulation Results Discussion}
The simulated human trials involved running each (environment, method, $N$) combination with 6 seeds, 108 experiments total, using $\rounds=4$ rounds of $\batches=20$ trajectory labels. We present the average evaluation results using $M=6$ tasks per emotion along with standard errors in Fig. \ref{fig:sim_evals}.

To evaluate hypothesis (2), improving learning efficiency for a set of emotions compared to an approach that models them independently, we juxtapose our method with SEP. Across all metrics our method is able to reach a higher performance faster, the gap growing with $N$, supporting our hypothesis. With SEP each query is only informative for one of $N$ emotive models. By contrast, with our method each query is informative for the entire VAD space and, thus, every evaluation emotion to some extent.

To evaluate hypothesis (3), matching learning efficiency for a single emotion compared to an approach that gets feedback specific to that emotion for each query, we juxtapose our method with SEP-ALL. SEP-ALL gets emotion specific cost labels for all $N$ emotions with each query, as opposed to the generic VAD label our method receives. Yet across the board performance of our method matches SEP-ALL supporting our hypothesis again. It is not practical to go beyond a few evaluation emotions with SEP-ALL since labeling overhead scales linearly with respect to $N$, whereas our method has constant overhead enabling capture of the full span of emotions. Furthermore, even for small $N$ providing VAD values (or natural language) may be easier than real valued emotion specific costs. 

Ultimately, VAD provides an interpretable latent representation that allows efficient learning of the space of emotive style with performance no worse than if we targeted any specific target emotion. Fig.~\ref{fig:cassie_visual} visualizes some emotive styles Cassie learned from this experiment, demonstrating our method's ability to work with high DOF, complex robots.

\begin{figure*}
    \includegraphics[width=0.99\textwidth]{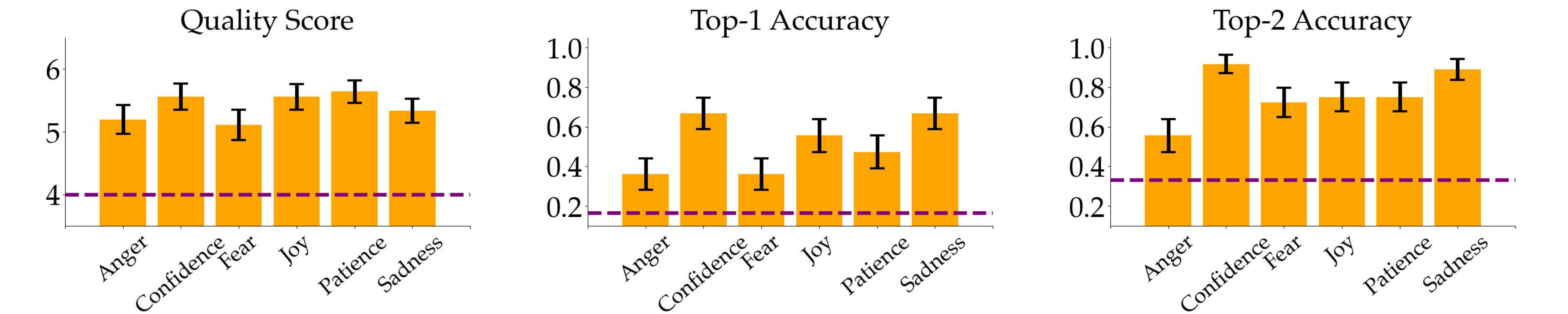}
    \centering
    \caption{Quality score, Top-1, and Top-2 accuracy and standard error for each evaluation emotion averaged across study users. Metrics significantly outperforms a random guess baseline, shown with the dashed lines, suggesting humans can indeed teach emotive motion with our method.}
    \label{fig:study_evals}
\end{figure*}

\begin{figure*}
    \includegraphics[width=0.99\textwidth]{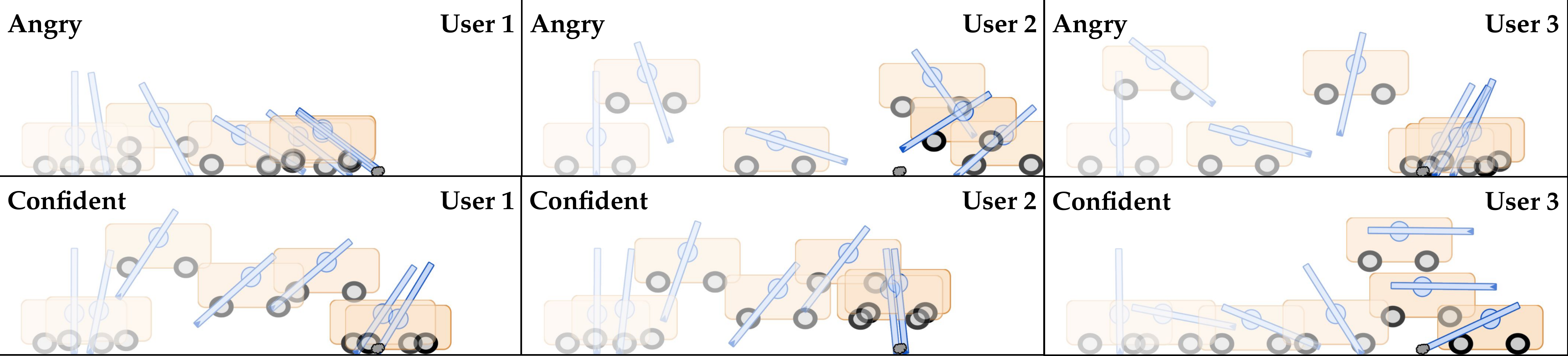}
    \centering
    \caption{VacuumBot collecting dust with style trained by three real users optimized for two target emotions on a single task. Judging by these motions, User 2 and 3's perception of anger involved greater speed, jumping, and arm movement than User 1. User 3's confidence had more arm movement but less jumping than User 1 and 2's.}
    \label{fig:study_variance}
\end{figure*}

\begin{figure*}
    \includegraphics[width=0.99\textwidth]{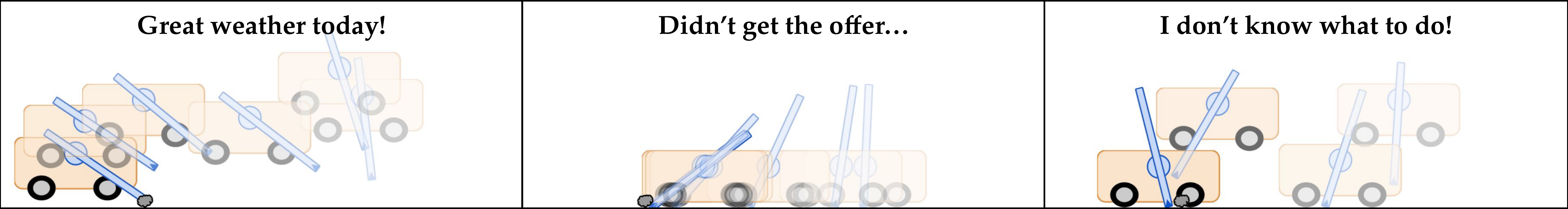}
    \centering
    \caption{Three trajectories for a single task where VacuumBot infers VAD from the displayed phrases and produces motions based on a study user's trained model. When the human expresses cheerfulness it gracefully hops to the goal. When the human expresses sorrow it slowly slouches its way there. When the human expresses fear it reflects that nervous energy.}
    \vskip -14pt
    \label{fig:study_NL}
\end{figure*}

\subsection{Real Human Study}
\label{sec:study}

As a result of the efficiency demonstrated in Sec.~\ref{sec:sims} it becomes feasible for real humans to use our system. In this section, we present a user study with VacuumBot aimed at testing hypothesis (1): the effectiveness of our method in teaching emotive style that is recognizable to end users.

\subsubsection{Study Setup}
We recruited 12 participants (9 male, 3 female) between 20 and 27 years old. They were asked to provide emotion labels for $\rounds=2$ rounds of $\batches=20$ robot trajectories. We use a lower value for $\rounds$ here compared to the simulated experiments to emphasize the practicality of our approach. We found it was easier for humans to consistently label trajectories with VAD labels directly; however, as discussed earlier language may be used as well.

For the evaluation phase we used $N=6$ emotions with $M=3$ tasks each for all participants. To keep the overall study time shorter we did not perform intermediate evaluations, only evaluating after all labeling was complete. The labeling portion of the study took 30-40 minutes and the evaluation phase 20-25 minutes per participant.

\subsubsection{Study Results Discussion}
Fig.~\ref{fig:study_evals} compares human evaluation results to a random guess baseline. We break down results by each of the 6 evaluation emotions.

For every emotion and metric we ran t-tests at the 5\% significance level comparing performance to the random baseline. Each test indicated statistically significant improvement which is reflected by the standard errors in Fig.~\ref{fig:study_evals}. These results support hypothesis (1): our participants could teach the robot by labeling query trajectories in about 30 minutes, and during evaluation they perceived the robot's intended emotion at a rate significantly higher than random chance.

While the quality scores are relatively consistent across emotions the Top-1 accuracy varies a fair amount. This suggests the method is reliable in producing trajectories that are generally in the right direction but might not exactly line up with the intended one. In some use cases the former may be of primary importance and in others emotional precision may be equally important; however, it is reassuring to know that the robot will roughly align with the intended style even when it is not perfect in targeting the particular one.

A qualitative observation is that all users had their own personalized views of emotions. As demonstrated in Fig.~\ref{fig:study_variance}, robots trained by three different study users ended up with fairly different, yet justifiable behaviors for the same emotions and task. Furthermore, in Fig.~\ref{fig:study_NL} we showcase example motions for one user's robot generated based on the VAD of short phrases. This highlights how our method enables the robot to learn more than a finite set of emotions.  It learns an entire emotive space which it may index into to generate appropriately expressive behavior. By making the teaching process more accessible, our work takes an important step towards enabling anyone to teach robots nuanced behaviors without needing a technical foundation themselves.
\section{Discussion}

\subsection{Summary}
We introduced a method that enables robots to perform functional tasks in ways that are expressive of a wide range of emotions. After being taught how trajectories map to VAD the robot may include a cost function encouraging a target emotive style in task motion optimization. Natural language VAD inference enables the robot to decide target emotions while in use and may also substitute numerical VAD labels during training.
Our experiments suggest that learning the VAD space jointly, beyond enabling emotion generalization, is more efficient and practical than trying to model each target emotion separately. Furthermore, our experiments provide evidence that our method enables real humans to teach robots discernible, emotive style.

\subsection{Limitations and Future Work}
\label{sec:limits}
First we share some short term directions. our environments only presented robot locomotion tasks, albeit varying the start, goal, and obstacle locations, hence expanding the task space to include more diverse objectives (e.g. object manipulation) would be an interesting direction. Although we mentioned the possibility of using language in place of VAD for training we did not explicitly evaluate this option. After training we propose inferring target VAD from user sentences, and despite promising qualitative results in Fig.~\ref{fig:study_NL} more in depth analysis is required. It's unclear even whether the robot should alter user emotion or merely reflect it.

Now we higlight some long term challenges. There are existing solutions for producing physical trajectories with our style cost \cite{Li2020AnimatedCA}, but bringing the training procedure into the physical domain is more challenging. It removes the ability to easily reset the robot as we do in simulation to facilitate label collection. Another challenge is while VAD captures the three most important emotive directions, sometimes differing emotions have similar VAD. For example, \textit{fear} and \textit{disgust} both have low valence and dominance with high arousal. Future work would have to navigate these subtleties while preserving the efficiency of our learning process.

We are excited about our results and believe they make an important contribution towards the end-goal of making robots more expressive and enabling people to teach personalized emotive styles. We look forward to seeing robots operate alongside humans with control over their exhibited emotion.






{
\bibliographystyle{IEEEtran}
\bibliography{IEEEabrv,references}
}

\end{document}